\documentclass[hidelinks,10pt,conference,letterpaper]{IEEEtran}

\IEEEoverridecommandlockouts

\usepackage{etex}

\usepackage{multicol}

\usepackage{comment}
\usepackage{siunitx}
\usepackage{relsize}
\usepackage{ifthen}

\usepackage[caption=false]{subfig}

\usepackage{graphics} 
\usepackage{rotating}
\usepackage{color}
\usepackage{enumerate}
\usepackage[T1]{fontenc}
\usepackage{psfrag}
\usepackage{epsfig} 
\usepackage{booktabs}
\usepackage{graphicx,url}
\usepackage{multirow}
\usepackage{array}
\usepackage{latexsym}
\usepackage{amsfonts}
\usepackage{amsmath}
\usepackage{amssymb}
\usepackage{xstring}
\usepackage{multirow}
\usepackage{xcolor}
\usepackage{prettyref}
\usepackage{flexisym}
\usepackage{bigdelim}
\usepackage{breqn} 
\usepackage{listings}
\let\labelindent\relax
\usepackage{enumitem}
\usepackage{xspace}
\usepackage{bm}
\graphicspath{{./figures/}}
\usepackage{tikz}
\usetikzlibrary{matrix,calc}
\usepackage{tkz-tab}

\usepackage{lineno}

\usepackage{mdwlist}

\let\itemize\undefined
\makecompactlist{itemize}{stditemize}

\newrefformat{prob}{Problem\,\ref{#1}}
\newrefformat{def}{Definition\,\ref{#1}}
\newrefformat{sec}{Section\,\ref{#1}}
\newrefformat{sub}{Section\,\ref{#1}}
\newrefformat{prop}{Proposition\,\ref{#1}}
\newrefformat{app}{Appendix\,\ref{#1}}
\newrefformat{alg}{Algorithm\,\ref{#1}}
\newrefformat{cor}{Corollary\,\ref{#1}}
\newrefformat{thm}{Theorem\,\ref{#1}}
\newrefformat{lem}{Lemma\,\ref{#1}}
\newrefformat{fig}{Fig.\,\ref{#1}}
\newrefformat{tab}{Table\,\ref{#1}}

\newcommand{\bdmath}{\begin{dmath}}
\newcommand{\edmath}{\end{dmath}}
\newcommand{\beq}{\begin{equation}}
\newcommand{\eeq}{\end{equation}}
\newcommand{\bdm}{\begin{displaymath}}
\newcommand{\edm}{\end{displaymath}}
\newcommand{\bea}{\begin{eqnarray}}
\newcommand{\eea}{\end{eqnarray}}
\newcommand{\beal}{\beq \begin{array}{lll}}
\newcommand{\eeal}{\end{array} \eeq}
\newcommand{\beas}{\begin{eqnarray*}}
\newcommand{\eeas}{\end{eqnarray*}}
\newcommand{\ba}{\begin{array}}
\newcommand{\ea}{\end{array}}
\newcommand{\bit}{\begin{itemize}}
\newcommand{\eit}{\end{itemize}}
\newcommand{\ben}{\begin{enumerate}}
\newcommand{\een}{\end{enumerate}}

\newcommand{\calA}{{\cal A}}
\newcommand{\calB}{{\cal B}}
\newcommand{\calC}{{\cal C}}

\newcommand{\calI}{{\cal I}}

\newcommand{\calM}{{\cal M}}

\newcommand{\calR}{{\cal R}}
\newcommand{\calS}{{\cal S}}
\newcommand{\calT}{{\cal T}}

\newcommand{\calV}{{\cal V}}

\newcommand{\hide}[1]{}

\newcommand{\hiddenText}{{\color{gray} hidden text.}}
\newcommand{\hideWithText}[1]{\hiddenText}

\newcommand{\selectedTraj}{\calS}
\newcommand{\attack}{\calA}

\usepackage{float}

\usepackage{hyperref}
\usepackage{graphicx}
\usepackage{epstopdf}
\usepackage{epsfig}
\usepackage{pgfplotstable}
\usepackage{pgfplots}

\graphicspath{{./figures/},{../figures/},{../},{./code/}}

\usepackage{xr}
\usepackage{cite}
\usepackage{amsthm}
\newtheoremstyle{mystyle}
  {}
  {}
  {\itshape}
  {}
  {\bfseries}
  {.}
  { }
  {\thmname{#1}\thmnumber{ #2}\thmnote{ (#3)}}
\theoremstyle{mystyle}

\floatname{algorithm}{Algorithm}

\newtheorem{mydef}{Definition}

\newtheorem{mytheorem}{Theorem}

\newtheorem{myproblem}{Problem}

\usepackage{algorithm}
\usepackage{algcompatible}

\title{\huge{Resilient Active Target Tracking with Multiple Robots}}

\author{
Lifeng Zhou,$^{1}$ Vasileios Tzoumas,$^{2}$ 
George J.~Pappas,$^{3}$ Pratap Tokekar$^{1}$
\thanks{$^{1}$The authors are with the Department of Electrical and Computer Engineering, Virginia Tech, Blacksburg, VA 24061 USA (email: {\fontsize{8}{8}\selectfont\ttfamily\upshape \{lfzhou, tokekar\}@vt.edu}).}
\thanks{$^{2}$At the time the paper was written, the author  was with the Department of Electrical and Systems Engineering, University of Pennsylvania, Philadelphia, PA 19104-6228 USA. Currently, the author is with the Laboratory for Information \& Decision Systems (LIDS), and the Department of Aeronautics and Astronautics (AeroAstro), Massachusetts Institute of Technology, Cambridge, MA 02139 USA (email: {\fontsize{8}{8}\selectfont\ttfamily\upshape vtzoumas@mit.edu}).}
\thanks{$^{3}$The author is with the Department of Electrical and Systems Engineering, University of Pennsylvania, Philadelphia, PA 19104 USA (email: {\fontsize{8}{8}\selectfont\ttfamily\upshape pappagsg@seas.upenn.edu}).}
\thanks{This work was partially supported 
by the ARL DCIST CRA W911NF-17-2-0181 program, and by the NSF grants 1566247 and 1637915.}
}

\begin{document}

\maketitle


\begin{abstract}
The problem of target tracking with multiple robots consists of actively planning the motion of the robots to track the targets. A major challenge for practical deployments is to make the robots resilient to failures. In particular, robots may be attacked in adversarial scenarios, or their sensors may fail or get occluded.  In this paper, we introduce planning algorithms for multi-target tracking that are resilient to such failures. In general, resilient target tracking is computationally hard. Contrary to the case where there are no failures, no scalable approximation algorithms are known for resilient target tracking when the targets are indistinguishable, or unknown in number, or with unknown motion model. In this paper we provide the first such algorithm, that also has the following properties: First, it achieves maximal resiliency, since the algorithm is valid for any number of failures.  Second, it is scalable, as our algorithm terminates with the same running time as state-of-the-art algorithms for (non-resilient) target tracking. Third, it provides provable approximation bounds  on the tracking performance, since our algorithm guarantees a solution that is guaranteed to be close to the optimal.  We quantify our algorithm's approximation performance using a novel notion of curvature for monotone set functions subject to matroid constraints.  Finally, we demonstrate the efficacy of our algorithm through MATLAB and Gazebo simulations, and a sensitivity analysis; we focus on scenarios that involve a known number of distinguishable targets. 
\end{abstract}


\section{Introduction}\label{sec:Intro}

Tasks such as surveillance, exploration, and security often require the capability to detect, localize, and track targets within a prescribed area.  
For example, consider the tasks:
\begin{itemize}
\item (\textit{Surveillance}) Detect and localize invasive fish in an ecosystem;~\cite{tokekar2013tracking}
\item (\textit{Area monitoring}) Detect and localize trapped people in a burning building;~\cite{grocholsky2006cooperative}
\item (\textit{Patrolling}) Detect and localize adversarial agents that move in an urban environment.~\cite{zengin2007real}
\end{itemize}
These tasks can greatly benefit by the use of robots that act as mobile sensors.  Indeed, advancements in robotic mobility, sensing, and communication envision the deployment of collaborative robots to support target tracking~\cite{kumar2012opportunities}. The problem of planning the (joint) motion of robots for target tracking is known as \emph{multi-robot active target tracking} in the literature~\cite{atanasov2014information}. 
This is a challenging problem due to the fact that the targets may be mobile and whose motion model may only be partially known. The targets may even be moving adversarially. There may be a large number of targets (more than the number of robots), even unknown in number, and may be indistinguishable from each other. Nevertheless, a number of algorithms have been designed that ensure near-optimal tracking for all of the aforementioned scenarios~\cite{atanasov2014information,robin2016multi,spletzer2003dynamic,frew2003observer,schwager2011eyes,pierson2017intercepting,tokekar2014multi,dames2017detecting}.

\begin{figure}
\centering
\includegraphics[width=0.75\columnwidth]{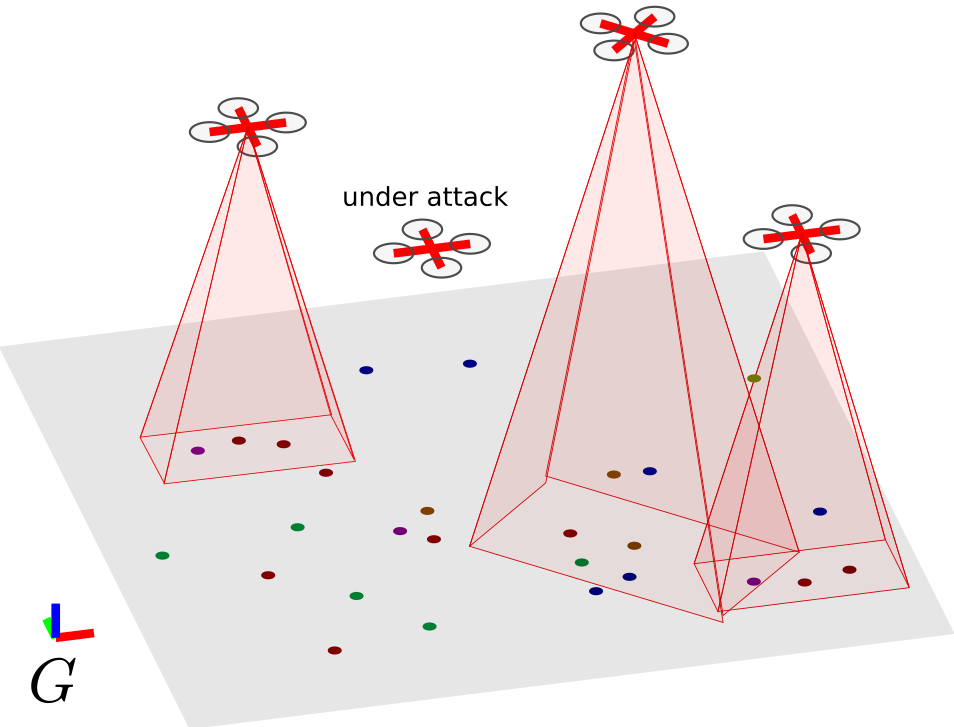}
\caption{Aerial robots mounted with down-facing cameras to track multiple targets ---depicted as dots--- on the ground.  The targets may be mobile; distinguishable or indistinguishable; known or unknown in number; or even with only partially known motion model. If a robot is under attack, its camera will be blocked. \label{fig:uav_target_tracking}}
\end{figure}

In this paper, we focus on scenarios where the robots operate in failure-prone or adversarial environments. In such scenarios the robots may be subject to attacks leading to robot failures~\cite{sless2014multi}, the robots' fields-of-view may become obstructed due to environmental hazards~\cite{gonzalez2002real}, or their sensors can fail completely~\cite{roumeliotis1998sensor} (see also Fig.~\ref{fig:uav_target_tracking}). 
We seek planning and coordination algorithms that are resilient to such failures and/or adversarial attacks.

In this paper, we introduce a problem of resilient target tracking that guards against worst-case robot failures even when the targets are indistinguishable, unknown in number, or even with unknown motion model.
Resilient target tracking is a computationally challenging problem since it needs to account for all possible robot and/or sensor failures, a problem of combinatorial complexity ---even in the presence of no failures, the problem is NP-hard~\cite{tokekar2014multi}. This computational challenge motivates the main goal in this paper: to provide a scalable and provably close-to-optimal approximation algorithm.  To this end, we capitalize on recent algorithmic results on \mbox{resilient optimization subject to matroids~\cite{tzoumas2018matroid}}, and present an approximation algorithm for resilient target tracking.

\medskip

\textbf{Contributions.} In this paper, we make the contributions:
\begin{itemize}
\item (\textit{Problem}) We formalize the problem of {resilient active target tracking} against \textit{worst-case} failures even in the presence of targets that are (possibly) indistinguishable,  unknown in number, or even of partially unknown motion model. 
This is the first work to formalize this problem. 
\item (\textit{Solution}) We develop the first algorithm for the problem, and prove it has the following properties:
\begin{itemize}
\item \textit{maximal resiliency}: the algorithm is valid for any number of robot and/or sensor failures;
\item \textit{minimal running time}: the algorithm terminates with the same running time as state-of-the-art algorithms for non-resilient target tracking;
\item \textit{provable approximation performance}: the algorithm ensures a close-to-optimal solution for any target tracking objective function that is monotone and submodular (submodularity is a diminishing returns property~\cite{fisher1978analysis}). Examples of such functions are the expected number of detected targets at a prescribed time and the mutual information between the predicted targets' location and the robots' sensor measurements~\cite{dames2017detecting}.  

\end{itemize}
\item (\textit{Empirical Evaluation}) We demonstrate with MATLAB and Gazebo simulations  both the necessity for resilient target tracking against robot failures, and the efficacy and robustness of our approach.  To this end, we focus on scenarios that involve distinguishable targets (known in number), and also conduct sensitivity analysis against non-worst-case attacks (random and greedy attacks).
\end{itemize}

Overall, in this paper we go beyond non-resilient target tracking~\cite{atanasov2014information,robin2016multi,spletzer2003dynamic,frew2003observer,schwager2011eyes,pierson2017intercepting,tokekar2014multi,dames2017detecting} by proposing {resilient} target tracking; and beyond resilient tracking with distinguishable and known targets~\cite{schlotfeldt2018resilient} by proposing resilient tracking with targets that are (possibly) {indistinguishable},  and/or unknown.

\medskip

\textbf{Organization of rest of the paper.} Section~\ref{sec:problemStatement} formulates resilient target tracking (Problem~1). Section~\ref{sec:alg}
presents the first scalable algorithm for Problem
1. Section~\ref{sec:performance_analysis} presents the main result in this paper: the scalability and performance guarantees of the proposed algorithm. Section~\ref{sec:exp} presents MATLAB and Gazebo simulations. Section~\ref{sec:con}
concludes~the~paper. 

\medskip

\textbf{Notation.} Calligraphic fonts denote sets (e.g., $\calA$).  Given a set $\calA$, $2^{\calA}$ denotes the power set of $\calA$; $|\calA|$ denotes $\calA$'s cardinality; given another set $\calB$, the set  $\calA\setminus\calB$ denotes the set of elements in $\calA$ that are not in~$\calB$. Given a set $\mathcal{V}$, a set function $f:2^\calV\mapsto \mathbb{R}$, and an element $x\in \mathcal{V}$,  $f(x)$ is a shorthand that denotes $f(\{x\})$.


\section{Problem Formulation} 
\label{sec:problemStatement}

We formalize the problem of resilient multi-target tracking. In particular, the problem 
consists of planning the motion of the robots to optimally track targets despite robotic/sensor failures.
The optimality of tracking is captured by an objective function such as the expected
number of detected targets or the reduction in the uncertainty of the targets' positions.

\subsection{Framework}\label{subsec:framework}

\paragraph*{Attacks} We assume that the maximum number of robotic/sensor failures are known, and denote it by~$\alpha$.\footnote{Henceforth, we refer to robotic and sensor failures interchangeably.} At any time at most $\alpha$ robots/sensor may fail. In addition, without loss of generality, the set of robots that fail may vary over time. A robot that fails at time $t$ may be active at another time~$t'$\!. 

The rest of our problem formulation, e.g., assumptions about the targets, robots, and the  objective function, follows the standard in the target tracking literature; see~\cite{dames2017detecting} and the references therein. 
Specifically:

\paragraph*{Targets} Targets exists in an area of interest (environment). The targets can be ground or aerial vehicles,  and can be mobile or immobile.  They can be distinguishable~\cite{tokekar2014multi} or indistinguishable~\cite{dames2017detecting}. Their number can be known~\cite{tokekar2014multi} or unknown~\cite{dames2017detecting}, fixed~\cite{tokekar2014multi} or time-varying~\cite{dames2017detecting}.  The target motion model can be known (e.g., a single integrator with known maximal speed~\cite{martinez2007motion}) or partially known; in the latter case, data-driven learning techniques may be employed~\cite{dames2017detecting}.

\paragraph*{Robots/sensors} We consider a team of mobile robots, and denote it by $\calR$. The team is tasked to track targets in an area of interest.  The robots can be ground or  aerial vehicles (e.g., quad-rotors).  We assume that the robots can communicate with each other at all times. 

The robots carry onboard sensors (e.g., cameras or lidars), which enable the team's tracking capability. In particular, each robot $r\in\calR$,  at every time $t$, receives measurements from targets detected in the field-of-view of its sensors. Additional measurements may be obtained from off-board sensors in the environment.  
Given the measurements and a target model, each robot employs a detector and trajectory estimator.  If the targets are distinguishable and their number is known, a Kalman or particle filter can be employed~\cite{thrun2005probabilistic}, whereas, if the targets are indistinguishable and their number is unknown, a Random Finite Sets (RFS) filter can be employed~\cite{dames2017detecting}.  
In both cases, the robots have only an estimate of the targets' true positions. The estimate is represented by a set of possible target locations in the environment. Given a target model, the robots propagate this set to obtain a predicted target position, by employing one of the above techniques. 

\begin{figure}
\centering
\includegraphics[width=0.65\columnwidth]{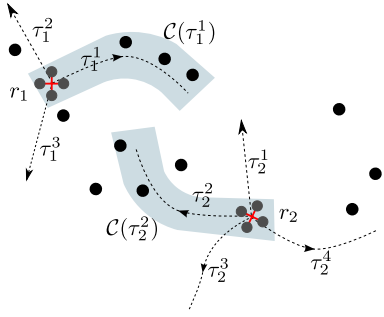}
\caption{Available robotic trajectories (denoted by $\tau_i^j$ for robot $r_i$) and their coverage region (denoted by $\calC(\tau_i^j)$ for robot $r_i$ that chooses trajectory $\tau_i^j$, and depicted as shaded regions).\label{fig:uav_tra_cover}}
\end{figure}

The robots can also move in the environment, and detect that way multiple targets per motion step (Fig.~\ref{fig:uav_tra_cover}).  The robot trajectory generation framework is as follows: We assume that the robots have perfect localization (e.g., using GPS).\footnote{The uncertainty in the robot's position can be incorporated in the uncertainty in the targets' estimates~\cite{tokekar2014multi}.}  
Time is divided into rounds of finite duration denoted by $T$ (without loss of generality, we assume it fixed).   At the beginning of each round, each robot generates a set of candidate trajectories, one of which will be followed in the current round. The trajectories can be generated by employing, for example, motion-space discretization~\cite{dames2017detecting} or spatial-sampling methods~\cite{green2010toward}.  
We denote the set of valid trajectories, for a round that starts at a time $t$, and for a robot $r\in\calR$, by $\calT_{r,t}$. We denote by $\calT_{\calR,t}$, the set of all robots' valid trajectories, i.e., $\calT_{\calR,t}\triangleq \cup_{r\in\calR}\calT_{r,t}$. 
Each trajectory in $\calT_{\calR,t}$ is interpolated to yield a sequence of robot poses where the robot will take a measurement. That is, each trajectory corresponds to a sequence of robot poses. 
Without loss of generality, we assume the number of poses to be fixed across rounds, robots, and trajectories. 

\paragraph*{Target tracking objective function} Given a round that starts at a time $t$, the utility of each trajectory in~$\calT_{\calR,t}$ is captured by an objective function $f$\!.
Two examples of $f$ are the following~\cite{dames2017detecting}: the expected number of detected targets 
in the current round (time interval from $t$ to $t+T$); and the mutual information between the
predicted location of the targets at time $t+T$ and the collected measurements in the current round.  Notably, both  functions are \textit{monotone} and \textit{submodular} in the choice of the robots' trajectories~\cite[Lemma~1 and Lemma~2]{dames2017detecting}.  Submodularity is a diminishing returns property~\cite{fisher1978analysis}; we provide its definition in Appendix~A of the full version of this paper, which is found at the authors' websites.  We henceforth focus on functions that satisfy these two properties. 

\subsection{Problem definition}

\begin{myproblem}[Resilient Multi-Target Tracking with Multiple Robots]\label{pro:res_track}
In reference to Section~\ref{subsec:framework}'s framework, consider: a set of targets;  a set $\calR$ of mobile robots/sensors; a division of time into rounds of finite duration. Moreover, consider the beginning of a round, and the corresponding set of valid robot trajectories $\calT_{\calR}\triangleq \cup_{r\in\calR}\calT_{r}$, where $\calT_{r}$ is the set of valid trajectories for the robot $r$. Finally, consider a target tracking objective function~$f$ that is monotone and submodular (e.g., the expected number of detected targets in the current round).  

The problem of \emph{resilient multi-target tracking with mobile robots} is to achieve a maximal value for $f$\!, by selecting the robot trajectories throughout the round, despite a worst-case failure of at most~$\alpha$ robots/sensors.  Formally:
\begin{align} \label{eq:res_track}
\begin{split}
\max_{\selectedTraj\subseteq \calT_{\calR}}\quad\min_{\attack\subseteq \selectedTraj} \quad f(\selectedTraj\setminus\attack):&\\
\medskip
\;\;\;|\mathcal{S}\cap \calT_{r}|= 1,~~~r\in\calR;&\\
 \;\;\;|\attack|\leq \alpha,&
\end{split}
\end{align}
where: $\calS$ is the set of selected trajectories for all robots; the constraint $|\mathcal{S}\cap \calT_{r}|\leq 1$ for each robot $r\in\calR$ represents the natural constraint that each robot $r$ can follow one trajectory;\footnote{This type of constraint is called a \emph{partition matroid} in the literature of combinatorial optimization~\cite{fisher1978analysis}.} 
and the constraint $|\calA|\leq \alpha$ captures the problem assumption that at most $\alpha$ robots/sensors can fail.
\end{myproblem}

Problem~\ref{pro:res_track} may be interpreted as a $2$-stage perfect information sequential game between two players~\cite[Chapter~4]{myerson2013game}, namely, a ``maximization'' player (who aims for optimal target tracking performance), and a ``minimization'' player (who aims to compromise the target tracking performance).  In particular,  the ``maximization'' player plays first by selecting the set~$\calS$, and, then, \emph{the ``minimization'' player observes $\calS$}, and plays second by selecting a worst-case attack/removal $\calA$ from $\calS$. Evidently, this is a stricter (worse) then version of the problem where the ``minimization'' player cannot observe $\calS$.

Problem~\ref{pro:res_track} goes beyond the traditional objective of  target tracking with mobile robots, by protecting (in a receding horizon fashion) the robots' motion plan against  failures.


\section{Algorithm for Problem~\ref{pro:res_track}}\label{sec:alg}

\begin{algorithm}[t]
\caption{Scalable algorithm for Problem~\ref{pro:res_track}.}
\begin{algorithmic}[1]
\REQUIRE Per Problem~\ref{pro:res_track}, Algorithm~\ref{alg:rob_sub_max} receives the inputs:
\begin{itemize}
\item set of robots $\calR$;
\item robot trajectories $\calT_r$, for all robot $r\in\calR$;
\item target tracking objective function $f$;
\item maximum number of failures $\alpha$.
\end{itemize}
\ENSURE Robots' trajectories $\mathcal{S}$.
\medskip

\STATE $\mathcal{S}_{1}\leftarrow\emptyset$;~~~$\mathcal{M}_{1}\leftarrow\emptyset$;~~~$\mathcal{S}_{2}\leftarrow\emptyset$;~~~$\mathcal{M}_{2}\leftarrow\emptyset$;\label{line:initiliaze}
\WHILE {$\mathcal{M}_{1}\neq \calT_\calR$}\label{line:begin_while_1}
\STATE $s\in \arg\max_{y \in \calT_\calR\setminus\mathcal{M}_{1}} f(y)$;\label{line:select_element_bait}
\IF {for all $r\in\calR$ it is $|(\mathcal{S}_{1}\cup\{s\})\cap \calT_{r}|\leq 1$, and $|\mathcal{S}_{1}\cup\{s\}|\leq \alpha$\!} \label{line:begin_if_1}
\STATE $\mathcal{S}_{1}\leftarrow\mathcal{S}_{1}\cup\{s\}$;\label{line:build_of_bait}
\ENDIF \label{line:end_if_1}
\STATE {$\mathcal{M}_{1}\leftarrow \mathcal{M}_{1}\cup \{s\}$}; \label{line:increase_removed_set_1}
\ENDWHILE \label{line:end_while_1}
\WHILE {$\calM_{2}\neq \calT_\calR\setminus \calS_1$} \label{line:begin_while_2} 
\STATE $s\in \arg\max_{y \in \mathcal{T}_\calR\setminus (\mathcal{S}_{1}\cup\calM_{2})}f(\calS_2\cup \{y\})-f(\calS_2)$; \label{line:greedy_selection}
\IF {for all $r\in\calR$ it is $|(\mathcal{S}_{1}\cup\mathcal{S}_{2}\cup\{s\})\cap \calT_r|\leq 1$} \label{line:begin_if_2}
\STATE $\mathcal{S}_{2}\leftarrow\mathcal{S}_{2}\cup\{s\}$;\label{line:build_of_greedy}
\ENDIF \label{line:end_if_2}
\STATE {$\calM_{2}\leftarrow \calM_{2}\cup \{s\}$}; \label{line:increase_removed_set_2}
\ENDWHILE \label{line:end_while_2}
\STATE $\mathcal{S}\leftarrow \mathcal{S}_{1} \cup \mathcal{S}_{2}$; \label{line:selection}
\end{algorithmic}\label{alg:rob_sub_max}
\end{algorithm}

We present the first scalable algorithm for Problem~\ref{pro:res_track}, by capitalizing on the algorithmic results in~\cite{tzoumas2018matroid}.
The pseudo-code of the algorithm is described in Algorithm~\ref{alg:rob_sub_max}.

\subsection{Intuition behind Algorithm~\ref{alg:rob_sub_max}}\label{subsec:intuition}

Problem~\ref{pro:res_track} selects trajectories for all robots, denoted by the set $\calS$ in eq.~\eqref{eq:res_track}, so to maximize the value of the objective function $f$ despite that $\calS$ can incur a removal $\calA$ of $\alpha$ elements due to robotic failures.
In~this~context, Algorithm~\ref{alg:rob_sub_max} aims to maximize $f$ by constructing $\calS$ as the union of two sets, the $\calS_{1}$ and~$\calS_{2}$ (line~\ref{line:selection}), whose role we describe below. 
\setcounter{paragraph}{0} 
\paragraph*{Set $\calS_{1}$ approximates a worst-case set removal from~$\calS$}  Algorithm~\ref{alg:rob_sub_max} aims with the trajectory set $\calS_{1}$  to capture a worst-case removal of $\alpha$ trajectories among the trajectories Algorithm~\ref{alg:rob_sub_max} will select in~$\calS$. Equivalently,~$\calS_{1}$ is aimed to act as a ``bait'' to an attacker that selects to remove the \textit{best}~$\alpha$ trajectories from~$\calS$  (\textit{best} with respect to the trajectories' contribution towards maximizing the function $f$). However, the problem of selecting the \textit{best} trajectories in~$\calT_\calR$ per Problem~\ref{pro:res_track} is combinatorial and, in general, intractable~\cite{hochbaum1998analysis}. 
For this reason, Algorithm~\ref{alg:rob_sub_max} aims to \textit{approximate} the best set of~$\alpha$ trajectories, by letting~$\calS_{1}$ be the trajectories with the largest contributions to the value of~$f$ (lines~\ref{line:select_element_bait}). In addition, since $\calS$ needs to satisfy the constraint that each robot $r\in\calR$ can be assigned one trajectory, Algorithm~\ref{alg:rob_sub_max} constructs $\calS_1$ so that not only $|\calS_1|\leq \alpha$ but also $|\mathcal{S}_{1}\cap \calT_{r}|\leq 1$ for all $r\in\calR$ (lines~\ref{line:begin_if_1}-\ref{line:end_if_1}).

\paragraph*{Set $\calS_{2}$ is such that the set $\calS_{1}\cup \calS_{2}$ approximates solution to Problem~\ref{pro:res_track}}
Assuming~$\calS_{1}$ is the set to be removed from Algorithm~\ref{alg:rob_sub_max}'s selection~$\calS$,
Algorithm~\ref{alg:rob_sub_max} needs to select a trajectory set $\calS_{2}$ to complete the construction of~$\calS$. In~particular, for $\calS=\calS_{1}\cup \calS_{2}$ to be a solution to Problem~\ref{pro:res_track}, 
Algorithm~\ref{alg:rob_sub_max} needs to select $\calS_{2}$ as a \textit{best} set of trajectories from $\calT_\calR\setminus\calS_{1}$ subject to the natural constraint that one trajectory is assigned to each robot (lines~\ref{line:begin_if_2}-\ref{line:end_if_2}).  
Nevertheless, the problem of selecting a \textit{best} set 
of elements subject to such a constraint is combinatorial and, in~general, intractable~\cite{hochbaum1998analysis}.  Hence, Algorithm~\ref{alg:rob_sub_max} aims to \textit{approximate} such a best set, 
using the greedy procedure in lines~\ref{line:begin_while_2}-\ref{line:end_while_2}.  

Overall, Algorithm~\ref{alg:rob_sub_max} constructs the sets $\calS_{1}$ and $\calS_{2}$ to approximate with their union $\calS$ an optimal solution to Problem~\ref{pro:res_track}.

We next describe the steps in Algorithm~\ref{alg:rob_sub_max} in more detail.

\subsection{Description of steps in Algorithm~\ref{alg:rob_sub_max}}

Algorithm~\ref{alg:rob_sub_max} executes four steps:
\setcounter{paragraph}{0}
\paragraph{Initialization (line~\ref{line:initiliaze})} Algorithm~\ref{alg:rob_sub_max} defines 4 sets, the $\calS_{1}$, $\calM_1$, $\calS_2$, and $\calM_2$, and initializes each of them with the empty set (line~\ref{line:initiliaze}). \textit{The~purpose of $\calS_{1}$ and $\calS_2$} is to construct the set $\calS$.
Specifically, 
the union of $\calS_{1}$ and~$\calS_2$ constructs~$\calS$ by the end of Algorithm~\ref{alg:rob_sub_max} (line~\ref{line:selection}).  \textit{The~purpose of $\calM_{1}$ and of $\calM_2$} is to support the construction of $\calS_{1}$ and~$\calS_2$. During the construction of $\calS_1$, Algorithm~\ref{alg:rob_sub_max} stores in $\calM_1$ the trajectories in $\calT_\calR$ that have either been included already or cannot be included in $\calS_1$ (line~\ref{line:increase_removed_set_1}); that way, Algorithm~\ref{alg:rob_sub_max} keeps track of which trajectories remain to be checked if they could be added in $\calS_1$ (line~\ref{line:build_of_bait}). During the construction of~$\calS_2$, Algorithm~\ref{alg:rob_sub_max} stores in $\calM_2$ the trajectories of $\calT_\calR\setminus\calS_1$ that have either been included already or cannot be included in $\calS_2$ (line~\ref{line:increase_removed_set_2}); that way, Algorithm~\ref{alg:rob_sub_max} keeps track of which trajectories remain to be checked if they could be added in $\calS_2$ (line~\ref{line:build_of_greedy}).

\paragraph{Construction of set $\calS_{1}$ (lines~\ref{line:begin_while_1}-\ref{line:end_while_1})} Algorithm~\ref{alg:rob_sub_max} constructs $\calS_{1}$ sequentially by adding one trajectory  at a time from $\calT_\calR$ to $\calS_{1}$.  Specifically, $\calS_1$, being the ``bait'' set, is constructed such that it satisfies both the trajectory assignment  constraint (one trajectory per robot) and the failures cardinality constraint (line~\ref{line:begin_if_1}). Also, $\calS_1$ is constructed such that each trajectory $s\in \calT_\calR$ added in~$\calS_1$ achieves the highest value of $f(s)$ among all the trajectories in $\calT_\calR$ that have not been yet added in~$\calS_1$ and can be added in $\calS_1$ (line~\ref{line:build_of_bait}).

\paragraph{Construction of set $\calS_{2}$ (lines~\ref{line:begin_while_2}-\ref{line:end_while_2})} Algorithm~\ref{alg:rob_sub_max} constructs the set $\calS_{2}$ sequentially, by picking greedily trajectories from the set $\calT_\calR\setminus \calS_{1}$ such that $\calS_1\cup\calS_2$ satisfies the trajectory assignment constraint in Problem~\ref{pro:res_track} (one trajectory per robot).
Specifically, the greedy procedure in Algorithm~\ref{alg:rob_sub_max}'s ``while loop''  (lines~\ref{line:begin_while_2}-\ref{line:end_while_2}) selects a trajectory $y\in\mathcal{T}_\calR\setminus (\calS_{1}\cup\calM_2)$ to add in $\calS_{2}$ only if $y$ maximizes the value of  $f(\calS_2\cup \{y\})-f(\calS_2)$, where the set~$\calM_2$ stores the trajectories that either have already been added to $\calS_2$ or have been considered to be added to $\calS_2$ but they were not since the resultant set $\calS_1\cup\calS_2$ would not satisfy the trajectory assignment constraint.

\paragraph{Construction of set $\calS$ (line~\ref{line:selection})}
Algorithm~\ref{alg:rob_sub_max} constructs the set $\calS$ as the union of the previously constructed sets $\calS_{1}$ and~$\calS_2$ \mbox{(lines~\ref{line:selection}).}  

In sum, Algorithm~\ref{alg:rob_sub_max} proposes a trajectory assignment $\calS$ as solution to Problem~\ref{pro:res_track}. In particular, Algorithm~\ref{alg:rob_sub_max} constructs $\calS$ to withstand any compromising robotic/sensor failure.

\section{Performance Analysis of Algorithm~\ref{alg:rob_sub_max}}\label{sec:performance_analysis}
We quantify the performance of Algorithm~\ref{alg:rob_sub_max}, by bounding its running time, and its approximation performance.  To this end, we use the following notion of curvature for set functions.

\subsection{Constrained curvature of monotone submodular functions}\label{subsec:curv}

\begin{mydef}[Constrained curvature]\label{def:curvature}
Consider a set~$\mathcal{T}$\!, and a non-decreasing submodular set function $f:2^{\mathcal{T}}\mapsto\mathbb{R}$ such that (without loss of generality) for any element $s \in \mathcal{T}$\!, it is  $f(s)\neq 0$.  Moreover, consider a collection of subsets of $\calT$\!, denoted by $\calI$; e.g., $\calI$ represents admissible sets where $f$ can be evaluated at. Then, the \emph{constrained curvature} of $f$ over $\calI$~is: \begin{equation}\label{eq:curvature}
\nu_f(\calI)\triangleq 1-\min_{\calS\in\calI}\min_{s\in\calS}\frac{f(\mathcal{S})-f(\mathcal{S}\setminus\{s\})}{f(s)}.
\end{equation}
\end{mydef}

The curvature $\nu_f$ measures how far $f$ is from being additive. In particular, Definition~\ref{def:curvature} implies $0 \leq \nu_f \leq 1$: If $\nu_f=0$, then for all sets $\calS\in\calI$ it holds $f(\mathcal{S})=\sum_{s\in \mathcal{S}}f(s)$. In~contrast, if $\nu_f=1$, then there exist a set $\calS\in\calI$ and an element $s\in\calT$ such that $f(\calS)=f(\calS\setminus\{s\})$; that is, in the presence of $\calS\setminus\{s\}$, the element $s$~loses all its contribution to the value of $f(\calS)$.  Notably, Definition~\ref{def:curvature} adapts the notion of curvature discussed in~\cite{conforti1984curvature} to the case where the set $\calS$ is constrained in an $\calI$, instead of $\calS$ being able to be any subset of $\calT$\!.

For example, in reference to the target tracking framework of Section~\ref{sec:problemStatement}, consider the expected number of detected targets as a function of the robot trajectories.  Then, this function has curvature zero if each robot detects different targets from the rest of the robots. In contrast, it has curvature one if, for example, at least two robots by following their trajectories receive the exact same measurements.

\subsection{Performance Analysis for Algorithm~\ref{alg:rob_sub_max}} \label{subsec:performance_analysis}

\definecolor{OliveGreen}{rgb}{0,0.6,0}
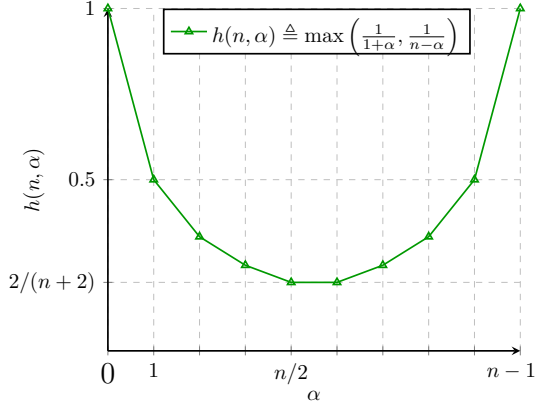
\begin{figure}[t]
 \begin{center}
 \begin{tikzpicture}[scale=0.8]
 \begin{axis}[
     axis lines = left,
     xtick = {0,1,...,9,10},
     xticklabels={\Large $0$,$1$,,,$n/2$,,,,,$\!\!\!\!\!n-1$},
     xlabel = $\alpha$,
     ytick = {0.2,0.5,1},
     yticklabels={${2}/(n+2)$,$0.5$,$1$},
     ylabel = {$h(n,\alpha)$},
     ymajorgrids=true,
     xmajorgrids=true,
     grid style=dashed,
     legend style={at={(0.88,1)}},
     ymin=0, ymax=1,
     line width=0.8pt,
 ]
 \addplot [
     domain=0:10, 
     samples=11, 
     color=OliveGreen,
     mark=triangle,
     ]
     {max(1/(1+x),1/(10-x))};
 \addlegendentry{$h(n,\alpha)\triangleq\max\left(\frac{1}{1+\alpha},\frac{1}{n-\alpha}\right)$}
 \end{axis}
 \end{tikzpicture}
 \caption{\small Given a natural number $n$,  plot of $h(n,\alpha)$ versus~$\alpha$.  Given  a finite~$n$, then $h(n,\alpha)$ is always non-zero, with minimum value $2/(n+2)$, and maximum value $1$.
 }\label{fig:bounds2}
 \end{center}
 \end{figure} 
 
\begin{mytheorem}[Performance of Algorithm~\ref{alg:rob_sub_max}]\label{th:alg_rob_sub_max_performance}
Consider an instance of Problem~\ref{pro:res_track}, the notation therein, the notation in Algorithm~\ref{alg:rob_sub_max}, and the definitions:
\begin{itemize}
\item let the number $f^\star$ be the (optimal) value to Problem~\ref{pro:res_track};
\item given a set $\mathcal{S}$ as solution to  Problem~\ref{pro:res_track}, let  $\mathcal{A}^\star(\mathcal{S})$ be a worst-case set removal from $\mathcal{S}$, per Problem~\ref{pro:res_track}, that is: 
$\mathcal{A}^\star(\mathcal{S})\in\arg\underset{\;\mathcal{A}\subseteq \calS, |\calA|\leq \alpha}{\text{\emph{min}}}   \; \;f(\mathcal{S}\setminus \mathcal{A})$. 
Evidently, a removal from $\calS$ corresponds to a set of robot/sensor failures;

\item define $h(|\calR|,\alpha)\triangleq \max [1/(1+\alpha), 1/(|\calR|-\alpha)]$.\footnote{A plot of $h(|\calR|,\alpha)$ is found in Fig.~\ref{fig:bounds2}.}
\end{itemize}
Finally, without loss of generality, consider that $f(\emptyset)=0$.

The performance of Algorithm~\ref{alg:rob_sub_max} is bounded as follows:

\begin{enumerate}[leftmargin=*]
\item \emph{(Approximation performance)}~Algorithm~\ref{alg:rob_sub_max} returns a trajectory set $\calS$ such that each robot is assigned a single trajectory, and:
\begin{equation}\label{ineq:bound_sub}
\frac{f(\mathcal{S}\setminus \mathcal{A}^\star(\calS))}{f^\star}\geq 
\frac{\max\left[1-\nu_f(\calI),h(|\calR|,\alpha)\right]}{2},
\end{equation}
where $\calI$ is the collection of valid trajectory assignments to robots per Problem~\ref{pro:res_track} (one trajectory per robot), that is: $\calI\triangleq \{\calS:\calS\subseteq \calT_\calR, |\calS\cap \calT_r|=1 \text{ for all } r\in \calR\}$. 

\item \emph{(Running time)}~Algorithm~\ref{alg:rob_sub_max} constructs the trajectory set $\calS$ as a solution to Problem~\ref{pro:res_track} with $O(|\calT_\calR|^2)$ evaluations of~$f$\!. 
\end{enumerate}
\end{mytheorem}

The proof of Theorem~\ref{th:alg_rob_sub_max_performance} follows the steps the proof of~\cite[Theorem~1]{tzoumas2018matroid}, and due to page limitations it is omitted. 

\textbf{Provable approximation performance.} Theorem~\ref{th:alg_rob_sub_max_performance}  implies on the approximation performance of Algorithm~\ref{alg:rob_sub_max}:
\setcounter{paragraph}{0}
\paragraph*{Near-optimality} Algorithm~\ref{alg:rob_sub_max} guarantees a value finitely close to the optimal,
for any monotone submodular objective function $f$: per ineq.~\eqref{ineq:bound_sub}, Algorithm~\ref{alg:rob_sub_max}'s approximation factor  is bounded by $h(|\calR|,\alpha)/2$, which is non-zero for any finite number of robots~$|\calR|$  (see also Fig.~\ref{fig:bounds2}).  Similarly, the~approximation factor is also bounded by $(1-\nu_f)/2$, which is also non-zero for any monotone submodular~$f$ with $\nu_f<1$.

\paragraph*{Approximation performance for no failures}
When the number of  failures is zero ($\alpha=0$), Algorithm~\ref{alg:rob_sub_max}'s approximation performance is the same as that of the state-of-the-art algorithms for (non-resilient) target tracking.  In~particular, these algorithms have approximation performance at least  $1/2$~\cite{tokekar2014multi,dames2017detecting};
at the same time, Algorithm~\ref{alg:rob_sub_max} also has performance at least $1/2$ for $\alpha=0$, since  $h(|\calR|,0)=1$ per ineq.~\eqref{ineq:bound_sub}.

\textbf{Minimal running time.}
Theorem~\ref{th:alg_rob_sub_max_performance} implies that Algorithm~\ref{alg:rob_sub_max}, even though it goes beyond the objective of (non-resilient) target tracking, 
has the same order of running time as state-of-the-art algorithms for (non-resilient) target tracking. In particular, these algorithms terminate with $O(|\calT_\calR|^2)$ evaluations of the function~$f$~\cite{tokekar2014multi,dames2017detecting}, and Algorithm~\ref{alg:rob_sub_max} also terminates with the same time.


\begin{figure}
\centering
\includegraphics[width=0.655\columnwidth]{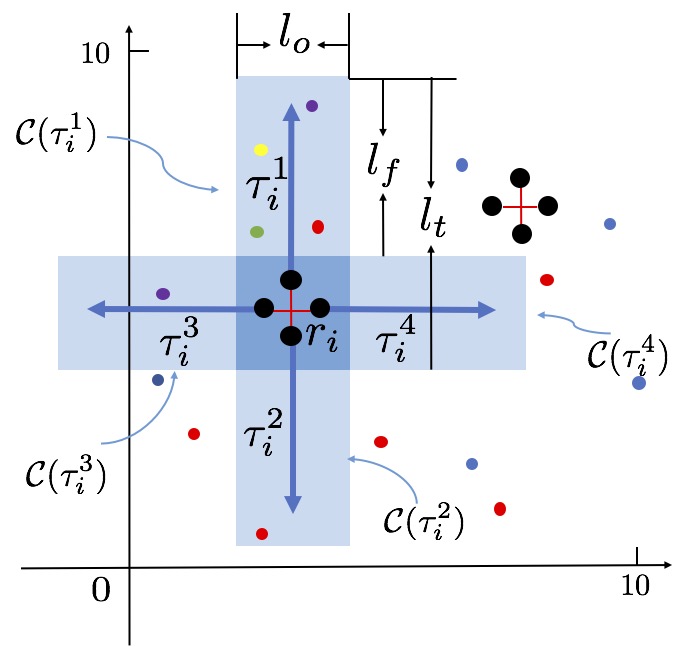}
\caption{MATLAB simulation setup: Each robot $r_i$ has 4 possible trajectories (forward, backward, left, and right, denoted by $\tau_i^j$ for $j=1,2,3,4$, respectively).  The tracking region of each trajectory is rectangular, is denoted by $\calC(\tau_i^j)$ for the trajectory $\tau_i^j$, and has the same dimensions across all 4 trajectories; in particular, the lengths $l_t$ and $l_o$ define the dimension of each rectangular region for each trajectory; and $l_f$ defines the fly length for the robot. We set $l_t=l_f+l_o$.\label{fig:rectangular_tracking}}
\end{figure}

\section{Numerical Evaluation}\label{sec:exp}
We present MATLAB and Gazebo evaluations of our algorithm (Algorithm~\ref{alg:rob_sub_max}) that demonstrate both the necessity for resilient target tracking and the benefits of our approach. In particular, both evaluations demonstrate: (i) the near-optimal performance of Algorithm~\ref{alg:rob_sub_max}, since the algorithm performs close to the brute-force algorithm (which is viable only in small-scale scenarios) and is superior to the greedy and random heuristics; and (ii) the superior robustness of Algorithm~\ref{alg:rob_sub_max} to scenarios where non-worst-case or even no attacks occur. Our MATLAB and Gazebo implementations are available online.\footnote{\url{https://github.com/raaslab/resilient_target_tracking.git}}

\begin{figure*}[htb]
\centering{
\hspace*{-5.5mm}
\subfloat[Optimal attack of $\alpha=3$ robots]{\includegraphics[width=0.515\columnwidth]{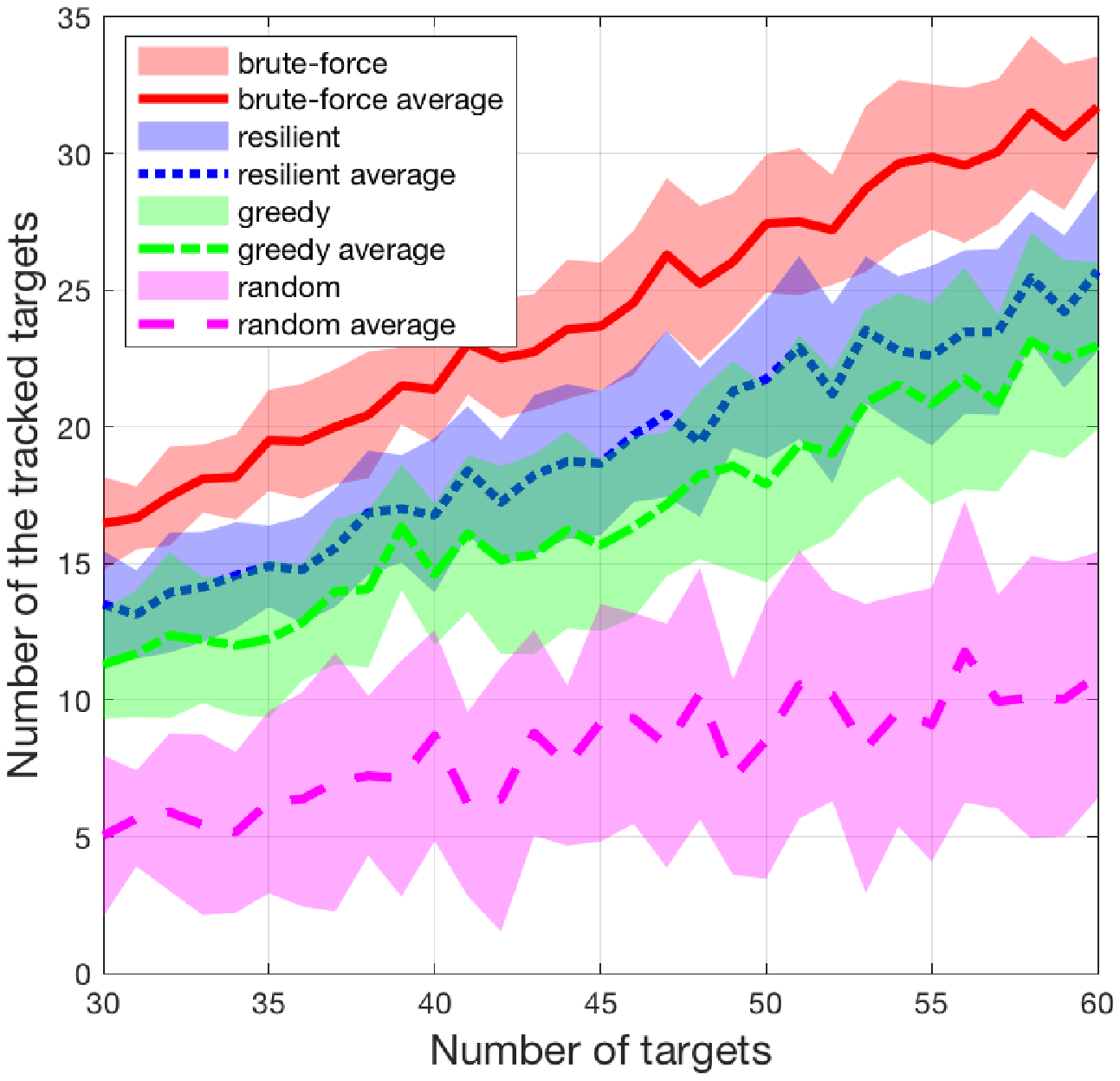}}
\subfloat[Optimal attack of $\alpha=4$ robots]{\includegraphics[width=0.515\columnwidth]{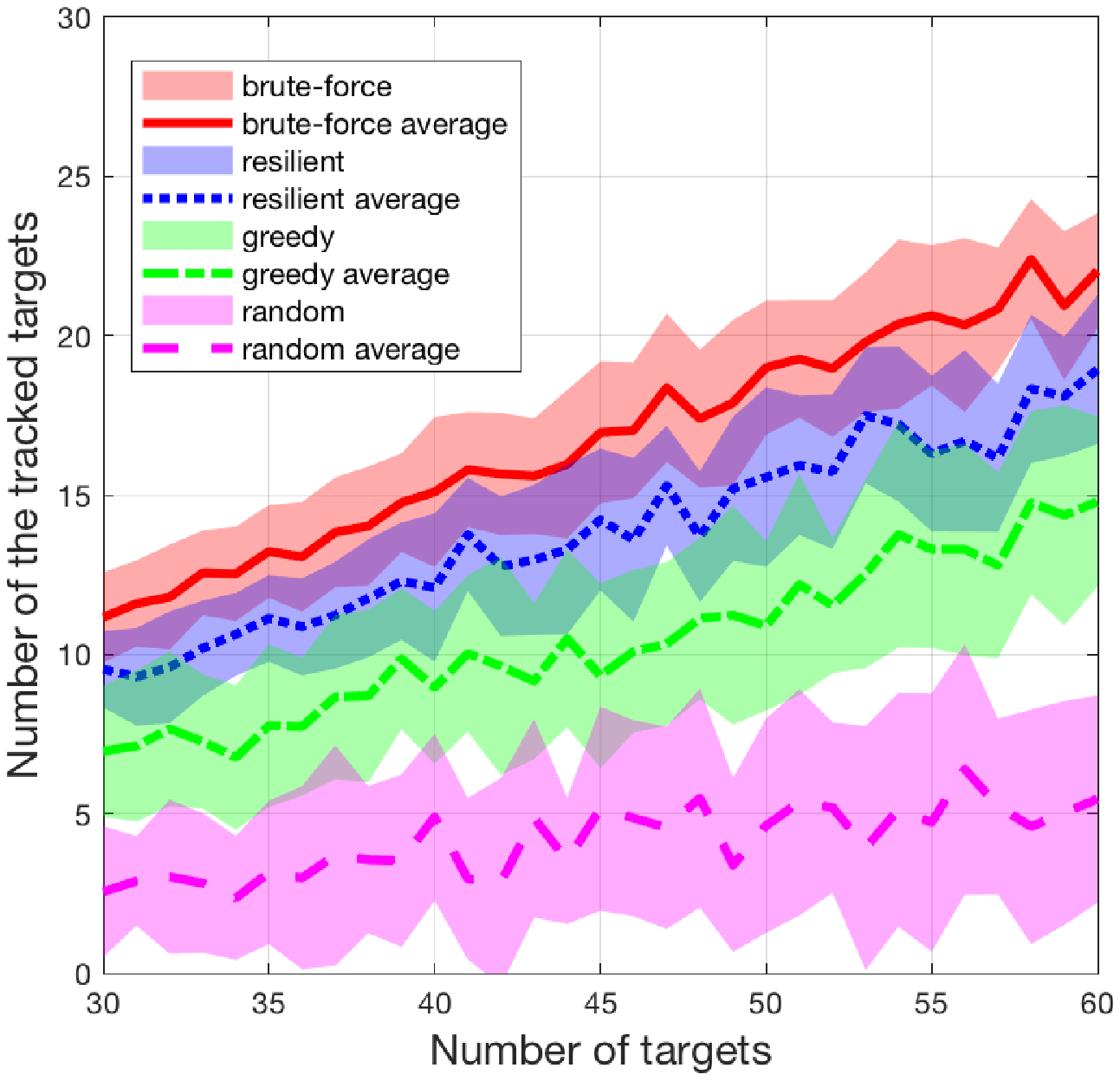}}
\subfloat[Greedy attack of $\alpha=3$ robots] {\includegraphics[width=0.515\columnwidth]{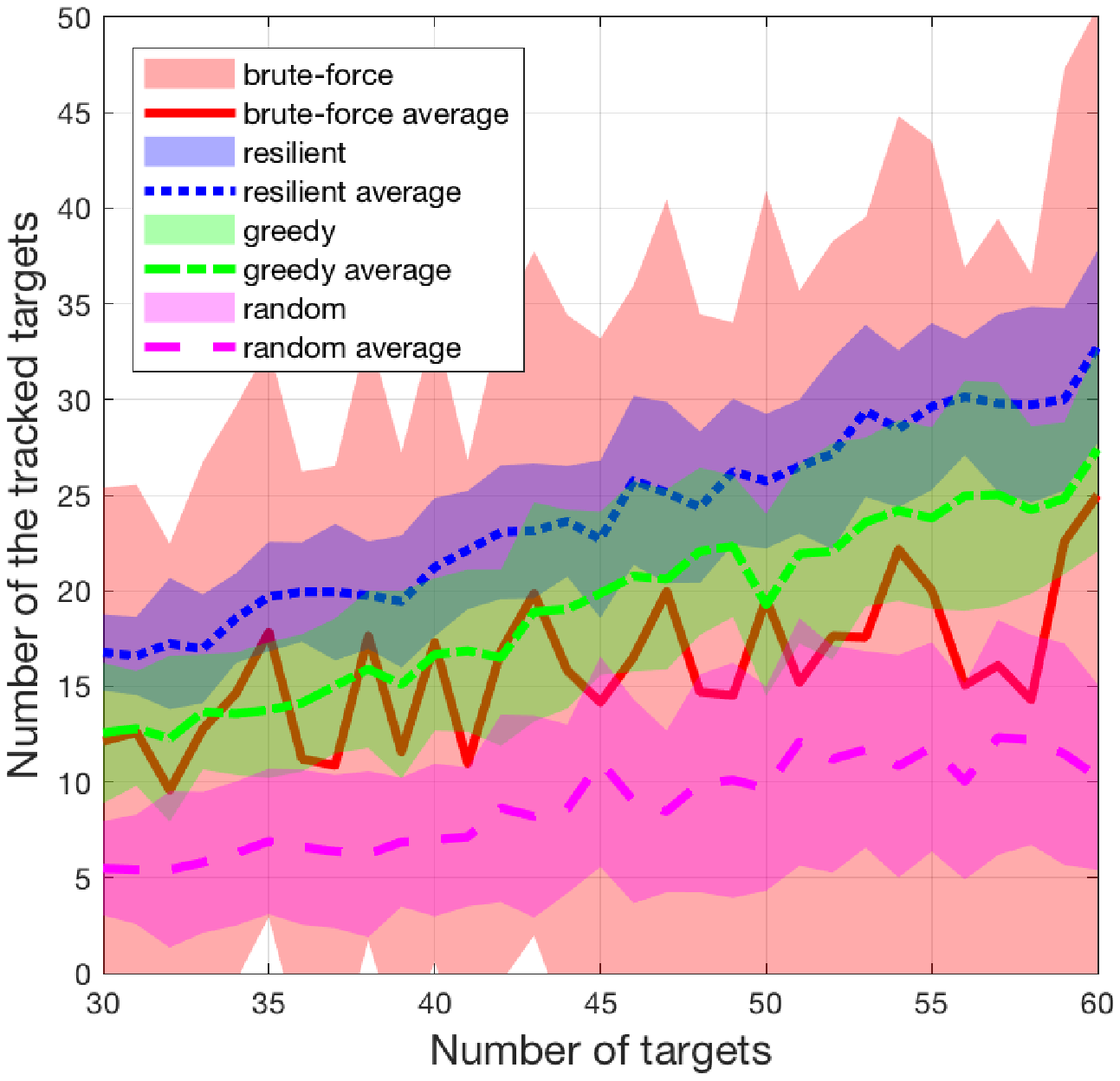}}
\subfloat[Random attack of $\alpha=3$ robots]{\includegraphics[width=0.515\columnwidth]{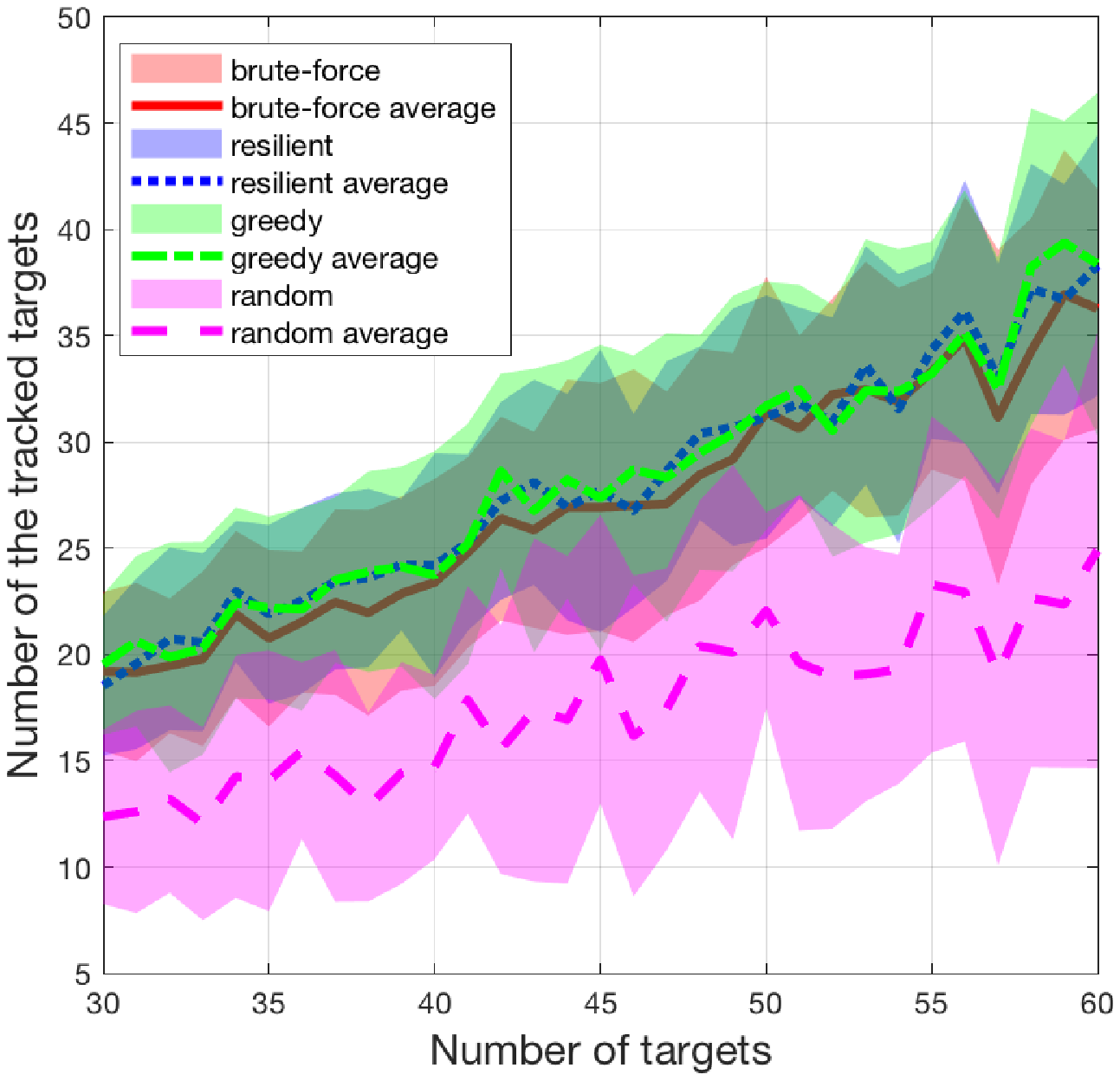}}
\caption{MATLAB evaluation results: Performance comparison (average and standard deviation over 30 trials) of Algorithm~\ref{alg:rob_sub_max} (called ``resilient'' in this figure) with the ``brute-force'' algorithm; the ``greedy'' algorithm; and the ``random'' algorithm. Performance is measured as the number of tracked targets, and is compared across 3 settings that differ on how an attacker would select to attack $\alpha$ robots so to minimize the number of tracked targets given the selected robot trajectories: in Fig.~\ref{fig:compare_matlab_opt}(a) and Fig.~\ref{fig:compare_matlab_opt}(b), the attacker uses a brute-force algorithm to find an optimal robot attack; in Fig.~\ref{fig:compare_matlab_opt}(c), the attacker uses the greedy algorithm in~\cite{fisher1978analysis}; and in Fig.~\ref{fig:compare_matlab_opt}(d), the attacker chooses randomly (uniformly across all robots).
\label{fig:compare_matlab_opt}}}
\end{figure*}

\textbf{Compared algorithms.} We compare Algorithm~\ref{alg:rob_sub_max} with three other algorithms. The
algorithms differ in how they select the robot trajectories. The first algorithm is an optimal, brute-force algorithm, and it attains the optimal value for Problem~\ref{pro:res_track}. Evidently, the brute-force approach is viable only when
the number of available robots is small. We refer to this algorithm by ``brute-force.'' The second algorithm is a greedy algorithm that ignores the possibility of robotic/sensor attacks, and picks greedily the robot trajectories per the algorithm proposed in~\cite{fisher1978analysis}; we refer to this algorithm by ``greedy.'' The third algorithm is a random algorithm that picks randomly (uniformly) the robot trajectories; we refer to this algorithm by ``random.'' Finally, we refer to Algorithm~\ref{alg:rob_sub_max} by ``resilient.''

\subsection{MATLAB evaluation over one step with static targets}

We study the effect of the number of targets and of the attack strategy by running the algorithms over random instances of Problem~1 for a single round (one-step time horizon).

\textbf{Simulation setup.} We consider 6 robots and a number of targets  $m$ that varies from 30 to 60. We set the number of attacks $\alpha$ equal to $3$ and $4$. A top view of the robots and targets is shown in Fig.~\ref{fig:rectangular_tracking}.  We assume that each robot $r_i\in \mathcal{R}$ flies on a fixed plane and has 4 trajectories: forward, backward, left, and right, denoted by $\tau_i^j$ for $j=1,2,3,4$, respectively. Each robot $r_i$ has a square field-of-view, centered at the planar position of robot $r_i$, and is illustrated in Fig.~\ref{fig:rectangular_tracking} by the darker blue square of dimension $l_o \times l_o$. Once each robot selects a trajectory, it flies a distance $l_f$ along that trajectory. Thus, each trajectory $\tau_{i}^{j}$ has a rectangular tracking region with length $l_t\triangleq l_f+l_o$ and width $l_o$; we set $l_t=10$ and $l_o = 3$ for all robots. 
For each number of targets $m=30,31,\ldots,60$, the planar positions of the robots and targets are randomly generated in the 2D space $[0,10]\times [0,10]\in\mathbb{R}^{2}$\!, across 30 trials. We consider that the robots have already available an estimate of the targets position. 
For each trial, all algorithms are executed with the same initialization, i.e., the same positions of targets and robots. All algorithms are executed for one round. 

The algorithms' performance is captured as the number of tracked targets given the selected robot trajectories.
We examine the performance across 3 settings that differ on how an attacker would select to attack $\alpha$ robots so to minimize the performance of the remaining robots: we first consider an attacker that uses a brute-force algorithm to find an optimal robot attack; this scenario is in agreement with the definition of Problem~\ref{pro:res_track}, where the attacks are indeed worst-case attacks.  Then, we consider an attacker that uses the greedy algorithm in~\cite{fisher1978analysis} to approximate an optimal robot attack; and finally, we consider an attacker that chooses randomly a robot attack (uniformly across all robots).  We examine the last two cases (Fig.~\ref{fig:compare_matlab_opt}(c) and Fig.~\ref{fig:compare_matlab_opt}(d)) as part of a sensitivity analysis of Algorithm~\ref{alg:rob_sub_max}'s performance against non-worst-case attacks.

\textbf{Results.} 
The comparison results are reported Fig.~\ref{fig:compare_matlab_opt}.  
The following observations from Fig.~\ref{fig:compare_matlab_opt} are due:

\textit{a)~~Close-to-optimality of Algorithm~\ref{alg:rob_sub_max}}: Algorithm~\ref{alg:rob_sub_max} is designed to guarantee superior performance in the presence of worst-case attacks; indeed, per Fig.~\ref{fig:compare_matlab_opt}(a) (and per Fig.~\ref{fig:compare_matlab_opt}(b)), Algorithm~\ref{alg:rob_sub_max} ---colored blue in  Fig.~\ref{fig:compare_matlab_opt}--- has on average superior performance to the greedy and random heuristics.  In particular, Algorithm~\ref{alg:rob_sub_max}'s performance is close to the optimal achieved by the brute force algorithm  (red in Fig.~\ref{fig:compare_matlab_opt}).

\textit{b)~~Robustness of Algorithm~\ref{alg:rob_sub_max}'s performance to non-worst-case attacks}: Although Algorithm~\ref{alg:rob_sub_max} is designed to guarantee superior performance for worst-case attacks, in practice, the attack of robots may not necessarily be the worst-case one. For example, from the perspective of an attacker, finding the optimal robot attack is also an NP-hard problem, since it constitutes a cardinality constrained submodular minimization problem~\cite{iyer2013fast}. It is therefore relevant to ask whether Algorithm~\ref{alg:rob_sub_max}, being an approximation algorithm,  will indeed have a better target tracking performance when the attacks are non-worst-case. 
By comparing  Fig.~\ref{fig:compare_matlab_opt}(a) with both Fig.~\ref{fig:compare_matlab_opt}(c) and Fig.~\ref{fig:compare_matlab_opt}(d), we observe that for each given number of targets (horizontal axes in each plot in Fig.~\ref{fig:compare_matlab_opt}) the performance of Algorithm~\ref{alg:rob_sub_max} increases for non-worst-case attacks. For example, for 30 targets, when the attack is worst-case (Fig.~\ref{fig:compare_matlab_opt}(a)) Algorithm~\ref{alg:rob_sub_max}  achieves 14 tracked targets, whereas: when the attack is greedy the performance increases to 17 (Fig.~\ref{fig:compare_matlab_opt}(c)). When the attacks are random, the performance increases to 18 (Fig.~\ref{fig:compare_matlab_opt}(d)). Overall, since Algorithm~\ref{alg:rob_sub_max} is designed to protect against worst-case attacks, it also protects at least equally well against non-worst-case attacks (as it would be expected). 
Importantly, that way an attacker is forced by Algorithm~\ref{alg:rob_sub_max} to deploy a worst-case attack, which is exactly the scenario that Algorithm~\ref{alg:rob_sub_max} guarantees protection from. That conclusion makes also irrelevant the observation that, for example, in the case of random attacks (Fig.~\ref{fig:compare_matlab_opt}(d)) both Algorithm~\ref{alg:rob_sub_max} and the greedy heuristic perform similarly. Notably, in the case of greedy attacks (Fig.~\ref{fig:compare_matlab_opt}(c)), Algorithm~\ref{alg:rob_sub_max} is again superior to both the greedy and the random heuristics.
In general, in the above simulation setup, Algorithm~\ref{alg:rob_sub_max} achieves a superior and close-to-optimal performance, and remains superior even against non-worst-case attacks.

\begin{figure}[t]
\centering{
\subfloat[Gazebo environment]
{\includegraphics[width=0.49\columnwidth]{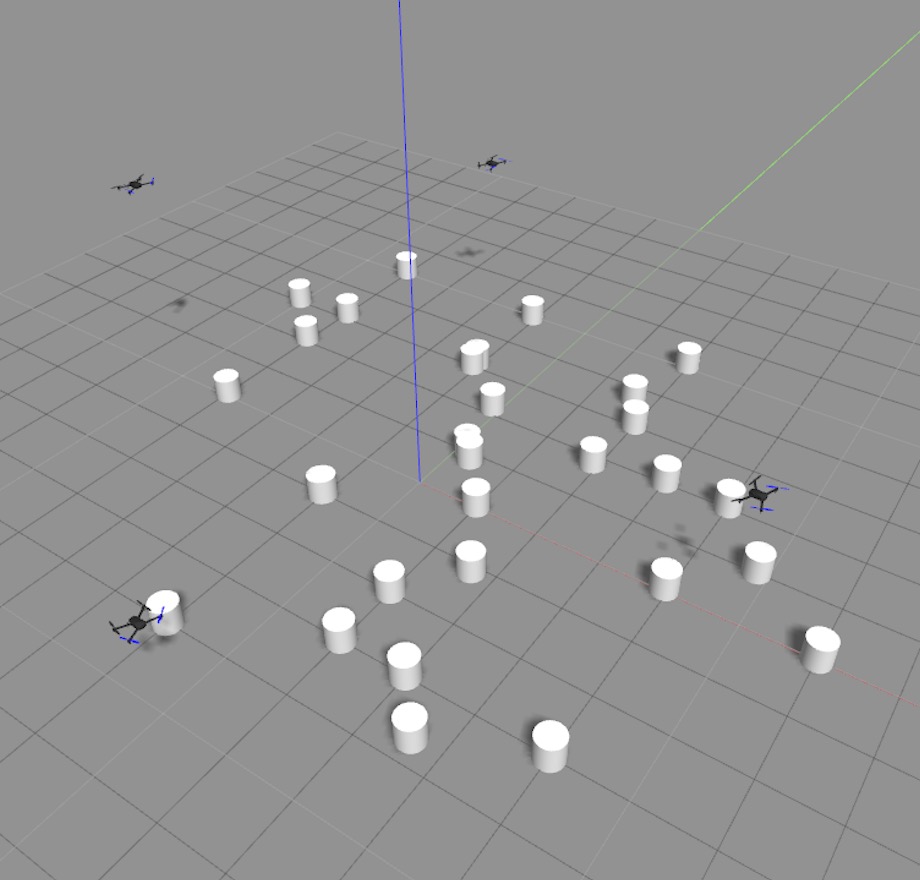}}~~
\subfloat[Rviz environment]
{\includegraphics[width=0.49\columnwidth]{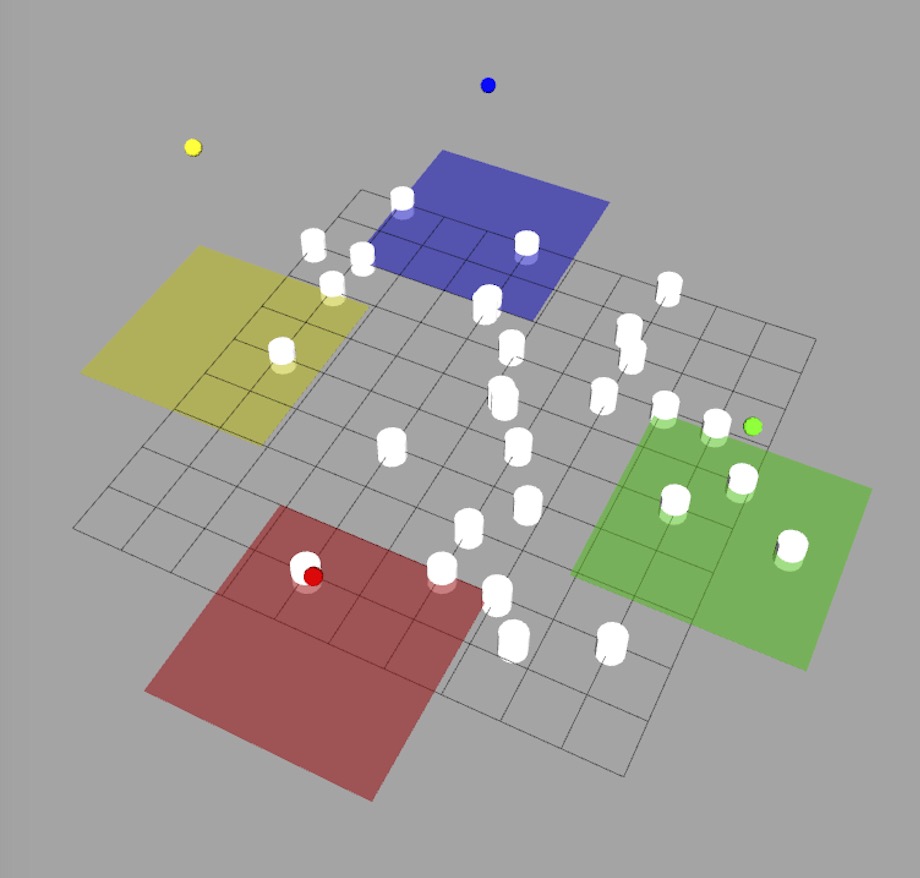}}
\caption{Gazebo simulation setup: 4 aerial robots and 30 ground mobile targets: (a) setup in Gazebo environment; and (b) setup in Rviz environment, where: each aerial robot is color-coded, and its coverage region is depicted with the same color.  The targets are depicted as white cylindrical markers. \label{fig:gazebo_rviz}}
}
\end{figure}

\subsection{Gazebo evaluation over multiple steps with mobile targets}

We study the effect of the number of targets and of the attack strategy by running the algorithms across multiple rounds (multi-step time horizon). That way, we take into account the kinematics and dynamics of the robots, as well as the fact that the kinematics and dynamics of the robot, the actual trajectories of the targets, and the sensing noise may force the robots to track fewer targets than expected.

\textbf{Simulation setup.} 
We consider a scenario of 4 aerial robots tasked to track 30 ground mobile targets (Fig.~\ref{fig:gazebo_rviz}(a)).  We set the number of attacks $\alpha$ equal to $2$. We also visualize the robots, their field-of-view, and the targets using the Rviz environment (Fig.~\ref{fig:gazebo_rviz}(b)): in particular, we visualize the robots as spherical markers, their field-of-views as colored areas with the same color as their corresponding robot, and the targets as white cylindrical markers. Similarly to the MATLAB simulation setup above, each robot has 4 trajectories (forward, backward, left, and right), and flies on a different fixed plane (to avoid collision with other robots).  Moreover, we set the tracking length $l_t = 6$ and width $l_o=3$ for all robots.  
We assume each target has the single integrator motion model
$$p_{t}^{j}(k+1) = p_{t}^{j}(k) + v_{t}^{j}(k),$$
where $p_{t}^{j}$ and $v_{t}^{j}$ denote the position and the velocity of target $j=1,\ldots,30$, respectively. The robots obtain noisy position measurements of all targets. 
They use a Kalman filter for updating the estimated position of the target at the next round. 
The targets' velocity is initialized to zero and is updated by using two consecutive measurements and the time interval of these two measurements, as follows: 
$$v_{t}^{j}(k') = (\tilde{p}_{t}^{j}(k') - \tilde{p}_{t}^{j}(k))/(k'-k).$$ where $\tilde{p}_{t}^{j}(k')$ and $\tilde{p}_{t}^{j}(k)$ are two consecutive position measurements of target $j$ at round $k'$ and round $k$ with $k'>k$.    

\begin{figure}[t]
\centering{
\subfloat[Performance comparison against worst-case attacks]{\includegraphics[width=0.7\columnwidth]{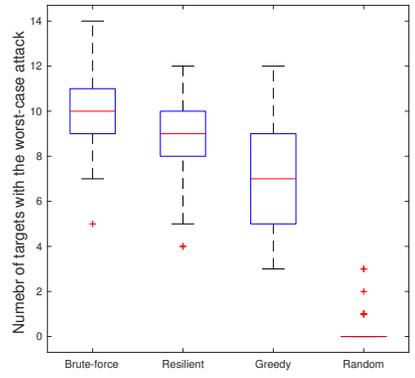}}\\
\vspace*{-3mm}
 \subfloat[Attack rate comparison]{\includegraphics[width=0.7\columnwidth]{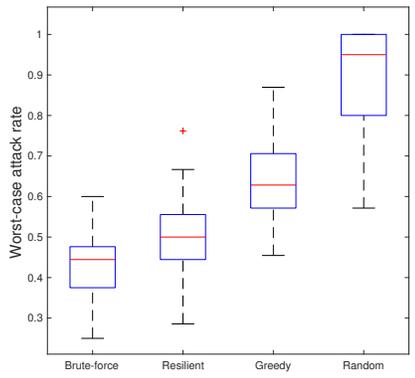}}
\caption{Gazebo evaluation results: Comparison (average and standard deviation across the 50 rounds) of Algorithm~\ref{alg:rob_sub_max} (called ``resilient'' in this figure) with the ``brute-force'' algorithm; the ``greedy'' algorithm; and the ``random'' algorithm. Performance is measured as the expected number of tracked targets. Fig.~\ref{fig:compare_gazebo}(b) compares the sensitivity of the algorithms' performance against the case where no attacks are present per the definition of the attack rate in eq.~\eqref{def:rem_rate} ---that is, the smaller the attack rate, the better.
\label{fig:compare_gazebo}}}
\end{figure}

For each algorithm (see ``Compared algorithms'' at the beginning of the section), at each round each robot selects one of its 4 trajectories.
Then, the robots fly a $l_f = 3$ distance along their selected trajectory.  
When an attack happens, we assume that the attacked robot's camera is turned-off; nevertheless, we assume that it can be active again at the next round, so that at each round the worst-case set of $\alpha$ robots is considered failed.
We repeat this process for 50 rounds. 

At each round, we capture the performance of each algorithm with the expected number of targets tracked. We first compare the algorithms with respect to the average and the standard deviation of the expected number of targets tracked. Moreover, we compare the sensitivity of the algorithms' performance against the case where no attacks are present: specifically, we compare the average and the standard deviation of their attack rate per round, which is defined ---for an algorithm that selects a set of trajectories $\calS$--- by
\begin{equation}\label{def:rem_rate}
\frac{f(\selectedTraj)-f(\selectedTraj\setminus\attack^{\star}(\mathcal{S}))}{f(\selectedTraj)},
\end{equation}
where $f(\selectedTraj)$ is the expected number of targets tracked in the presence of no attacks, and $f(\selectedTraj\setminus\attack^{\star}(\mathcal{S}))$ is the expected number of targets tracked in the presence of an optimal attack $\attack^{\star}(\mathcal{S})$. All in all, the above definition of attack rate captures how much worse is the performance of an algorithm in the presence of attacks than in the absence of attacks.
A video for this implementation is available online.\footnote{\url{https://youtu.be/KOeMHybX22M}}

\textbf{Results.} 
The comparison results are reported Fig.~\ref{fig:compare_gazebo}.  
The following observations from Fig.~\ref{fig:compare_gazebo} are due:

\textit{a)~~Close-to-optimality of Algorithm~\ref{alg:rob_sub_max}}:  Fig.~\ref{fig:compare_gazebo}(a) suggests that Algorithm~\ref{alg:rob_sub_max} has on average superior performance than the current heuristics (the greedy and the random).  In particular, Algorithm~\ref{alg:rob_sub_max}'s performance is close to the optimal, as it is achieved by a brute force algorithm.

\textit{b)~~Robustness of Algorithm~\ref{alg:rob_sub_max}'s performance to no-attacks}: Per Fig.~\ref{fig:compare_gazebo}(b), Algorithm~\ref{alg:rob_sub_max} exhibits superior attack rate than the greedy and random  heuristics, and as a result, when  for example the scenario ``at most $\alpha$ attacks per round'' and the scenario ``no attacks per round'' happen with equal probability, Algorithm~\ref{alg:rob_sub_max} still guarantees superior average performance.

All in all, in the above simulation setup, Algorithm~\ref{alg:rob_sub_max} achieves a superior and close-to-optimal performance, and remains superior even when no-attacks happen.

\section{Concluding Remarks \& Future Work}\label{sec:con}

We take the first steps to protect critical
target tracking tasks from robot failures (Problem 1). In particular, we provide
the first algorithm for Problem 1, and proved its
guaranteed performance against any number of failures, and even for targets that are indistinguishable and/or unknown. We demonstrate the need for resilient target tracking and the robustness of our algorithm with MATLAB and Gazebo evaluations.

This work opens a number of avenues for future research, both theoretical and experimental. Future theoretical work in theory
includes the {decentralized} design of the robots' motion plan. Moreover, online extensions of Algorithm~1, that guarantee near-optimality across multiple rounds, are natural next steps.
Future experimental work includes real-world  testing
of our resilient target tracking framework in the context of practical applications of surveillance and patrolling.

\bibliographystyle{IEEEtran}
\bibliography{references}

\appendices

\section{Monotonicity and Submodularity}\label{app:monotone_submodular}

We define monotonicity and submodularity, and discuss them within the target tracking  framework of Section~\ref{sec:problemStatement}. 

\begin{mydef}[Monotonicity~{\cite{nemhauser78analysis}}]\label{def:mon}
Consider a finite ground set~$\calT$\!. Then, a set function $f:2^\calT\mapsto \mathbb{R}$ is \emph{non-decreasing} if and only if for any sets $\mathcal{S}\subseteq \mathcal{S}'\subseteq\calT$\!, it holds $f(\mathcal{S})\leq f(\mathcal{S}')$.
\end{mydef}

For example, the expected number of detected targets is a non-decreasing function in the choice of robot trajectories: as the number of robots participating in the target tracking increases, the expected number of detected targets also increases.

\begin{mydef}[Submodularity~{\cite[Proposition 2.1]{nemhauser78analysis}}]\label{def:sub}
Consider any finite set $\calT$\!.  Then, the set function $f:2^\calT\mapsto \mathbb{R}$ is \emph{submodular} if and only if
for any sets $\mathcal{S}\subseteq \mathcal{S}'\subseteq\calT$\!, and any element $s\in \calT$\!, it {holds}  
$f(\mathcal{S}\cup \{s\})\!-\!f(\mathcal{S})\geq f({\mathcal{S}'}\cup \{s\})\!-\!f({\mathcal{S}'})$.
\end{mydef}

Definition~\ref{def:sub} implies that a set function $f$ is submodular if and only if it satisfies the following diminishing returns property:
for any set $\mathcal{S}\subseteq \calT$\!, and any element $s\in \calT$\!, the marginal gain $f(\mathcal{S}\cup \{s\})-g(\mathcal{S})$ is~non-increasing. 

For example, the expected number of detected targets is submodular in the choice of robot trajectories~\cite[Lemma~2]{dames2017detecting}: in the presence of more robots, the addition of a robot has a smaller effect on increasing the number of detected targets.

\end{document}